# Autoencoder, Principal Component Analysis and Support Vector Regression for Data Imputation

Vukosi N. Marivate*. Fulufhelo V. Nelwamodo**
Tshilidzi Marwala***

*School of Electrical and Information Engineering, University of the Witwatersrand,
Johannesburg, 2050, South Africa
*(e-mail: vukosi.marivate@students.wits.ac.za).
**(e-mail: f.nelwamondo@ee.wits.ac.za).
*** (e-mail: tshilidzi.mawala@wits.ac.za).*

Abstract: Data collection often results in records that have missing values or variables. This investigation compares 3 different data imputation models and identifies their merits by using accuracy measures. Autoencoder Neural Networks, Principal components and Support Vector regression are used for prediction and combined with a genetic algorithm to then impute missing variables. The use of PCA improves the overall performance of the autoencoder network while the use of support vector regression shows promising potential for future investigation. Accuracies of up to 97.4 % on imputation of some of the variables were achieved.

## 1. INTRODUCTION

Acquired immunodeficiency syndrome (AIDS) is a collection of symptoms and infections resulting from the specific damage to the immune system caused by the human immunodeficiency virus (HIV) in humans (Marx, 1982). South Africa has seen an increase in HIV infection rates in recent years as well as having the highest number of people living with the virus. This results from the high prevalence rate as well as resulting deaths from AIDS (Department of Health, 2000). Research into the field is thus strong and ongoing so as to try to identify ways into dealing with virus in certain areas. Thus demographic data is used often to class people living with aids and how they are affected. Thus proper data collection needs to be done so as to understand where and how the virus is spreading. This will give more insight into ways in which education and awareness can be used to equip the South African population. By being able to identify factors that deem certain people or populations in higher risk, the government can then deploy strategies and plans within those areas so as to help the people.

The problem with data collection in surveys is that is suffers from information loss. This can result from incorrect data entry or an unfilled field in a survey. This investigation explores the field of data imputation. The approach taken is to use regression models to model the interrelationships between data variables and then undertake a controlled and planned approximation of data using the regression model and an optimisation model. Data imputation using Auto Encoder Neural Networks as a regression model has been carried out by Abdella and Marwala (Mussa *et al*, 2005) and others (Leke *et al*, 2005) *(*Nelwamondo *et al,* 2007a*)* while other variations are available in literature including Expectation Maximisation *(*Nelwamondo *et al,* 2007a*)*, Rough Sets (Crossingham *et al*, 2005) *(*Nelwamondo *et al,* 2007b*)*, Decision Trees (Barcena *et al*, 2002). The use of Auto Encoder Networks comes with the price of computational complexity and a time trade-off as a disadvantage that is mostly cited for the use of other methods (Nelwamondo *et al*, 2007b), . The advantage of using Auto Encoder Networks it the high level of accuracy. The data used in this investigation is HIV demographic data collected from ante-natal clinics from around South Africa.

This report focuses on investigating the use of different regression methods that offer a glance into the data imputation world. The report first gives a background into missing data, neural networks and the other regression methods used. Secondly the data set to be used is introduced and explained. The methodology is given and then carried through. By the end of the report the results are given and then discussed.

## 2. BACKGROUND

*2.1 Missing Data*

Data collection forms the backbone of most projects and applications. To accurately use the data all information required must be available. Data collections suffer from missing values/data variables. This for example can be in the form of unfilled fields in a survey or data entry mistakes. Simply removing all entries concerned with the missing value is not always the best solution. There are three different types of missing data mechanisms as discussed by Little and Rubin (Little *et al*, 2000).

- Missing Completely at Random (**MCAR**) – This is when the probability of the missing value of a variable *x* is unrelated to itself or any other variables in the data set.

- Missing at Random (**MAR**) – This implies that the probability of missing data of a particular variable *x* depends on other variables but not itself

- Non-ignorable – This is when the missing value of variable *x* depends on itself even though other variables are known

Methods are needed to impute the missing data. There are numerous ways that have been used to impute missing data. The approach taken in this investigation is to use regression methods to find the inter-relationships between the data and then use the regression methods to verify the approximations that are made. The next subsections discuss the different regression methods used.

*2.2 Neural Networks*

Neural Networks are computational models that have the ability to learn and model systems. They have the ability to model non-linear systems (Bishop, 1995). The neural network architecture used is a multilayer perceptron network (Bishop, 1995) as shown in Fig. 1.

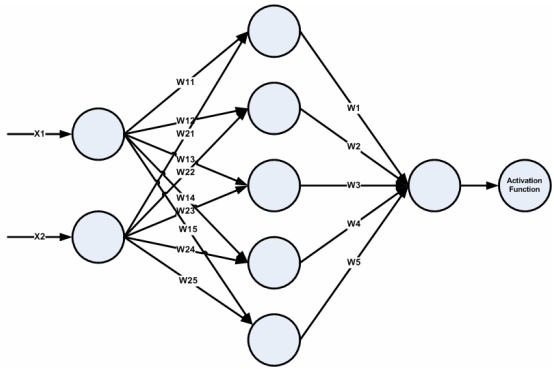

Fig. 1 MLP Neural Network

This has two layers of weights which connect the input layer to the output layer. The middle of the network is made up of a hidden layer. This layer can be made up of a different number of hidden nodes. This number has to be optimised so that the network can model systems better (Krose *et al*, 1996). An increase in hidden nodes translates into an increase in the complexity of the system. The output and the hidden nodes also have activation functions (Bishop, 1995). The general equation of a MLP neural network is shown below (1):

$$y_k = f_{outer}(\sum_{j=1}^{M} w_{kj}^{(2)} * f_{inner}(\sum_{i=1}^{d} w_{ji}^{(1)} + w_{j0}^{(1)}) + w_{k0}^{(2)}) \quad (1)$$

The activation function (*Fouter*) chosen for the project was linear. The inner activation (*Finner*) function chosen was the hyperbolic tangent function (tanh). This served to increase accuracy in regression (Krose *et al*, 1996). This function produced the best results during training. Thus the relation becomes (2):

$$y_k = \sum_{j=1}^{M} w_{kj}^{(2)} * \tanh(\sum_{i=1}^{d} w_{ji}^{(1)} + w_{j0}^{(1)}) + w_{k0}^{(2)} \quad (2)$$

For this project the Netlab (Nabney, 2001) MATLAB toolbox was utilised. The Netlab toolbox was used to implement the neural networks.

*2.3 Auto-encoder Networks*

Autoencoder/Auto Associative neural networks are neural networks that are trained to recall their inputs. Thus the number of inputs is equal to the number of outputs. Autoencoder neural networks have a bottleneck that results from the structure of the hidden nodes (Thompson *et al.*, 2002). There are less hidden nodes than input nodes. This results in a butterfly structure. The autoencoder network is preferred in recall applications as it can map linear and nonlinear relationships between all of the inputs. The autoencoder structure results in the compression of data into a smaller dimension and then decompressing into the output space. Autoencoders have been used in a number of applications including missing data imputation (Mussa & Marwala, 2005) (Nelwamondo *et al*, 2007a).

In this investigation an auto encoder networks was constructed using the MLP structure discussed in the previous subsection. The HIV data was fed into the network and the networks was trained to recall the inputs. Thus the structure is as in Fig. 2:

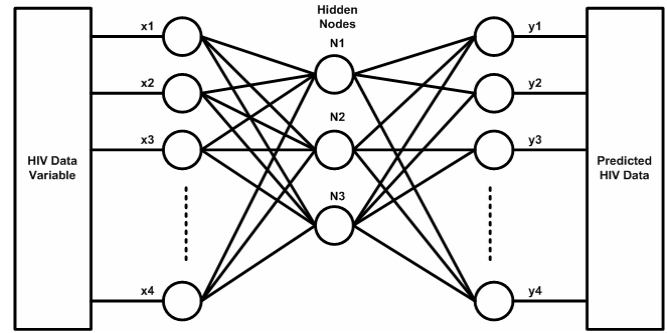

Fig. 2. Autoencoder Neural Network

*2.4 Support Vector Regression*

Support Vector Machines are a supervised learning method used mainly for classification. Support vector machines are classifiers derived from statistical learning theory and were first introduced by Vapnik (1998). They have also been extended to regression thus resulting in the term Support Vector Regression (SVR) (Gunn, 1997). In support vector regression the input *x* is mapped to a higher dimensional feature space $\Phi(x)$ in a non linear manner. This is depicted by (3) where *b* is the threshold for the support vector equation.

$$f(x) = (w \cdot \Phi(x)) + b \quad (3)$$

*w* and *b* are constants and can be estimated by reducing the empirical risk and a complexity term. The above equation is for a linear approximation of a function. $(w \cdot \Phi(x))$ describes the dot product between w and $\Phi(x)$. In (4) below the first term is the empirical risk and the second term represents the complexity.

$$R_{reg}[f] = R_{emp}[f] + \lambda \|w\|^2$$
$$= C \sum_{i=1}^{Z} (f(x_i) - y_i) + \lambda \|w\|^2 \quad (4)$$

The reduction of (4) is subject to the minimisation of the complexity as well as the optimisation of the regularisation parameter $\lambda$. The constant $\lambda>0$ determines the trade-off between the flatness of $f$ and the amount up to which deviations larger than $\varepsilon$ are tolerated. $C$ is the cost function. $Z$ is the number of records in the training set. By introducing $\varepsilon$ term then modelling non linear functions can be done. The non linear modelling can be at very high dimensions and can take long to compute solutions. To make the computation easier kernel functions are used (Gunn, 1997). There are numerous kernel functions and the one employed in the investigation is the Radial basis Function kernel. For an in depth tutorial on support vector machines for classification and regression see the tutorial by Gunn (1997). The least squares support vector toolbox was used for the investigation (Suykens et al, 2002).

*2.5 Principal Component Analysis*

Principal component analysis (PCA) (Shlens, 2005) is a statistical technique that is commonly used to find patterns in high dimensional data (Smith, 2002). The data can then be expressed in a way that highlights its similarities and differences. Another property is that, after finding the patterns in the data the data can then be compressed without much data loss. This is advantageous for ANNs as it will result in a reduction of the number of nodes needed, thus increasing computational speeds. A principal component analysis takes place in the following manner. First data is taken and the mean of each dimension is subtracted from the data. Secondly the covariance matrix of the data is then calculated. Thirdly the eigenvalues and eigenvectors of the covariance matrix are calculated. The highest eigenvalue corresponds to the eigenvector that is the principal component. This is where the notion of data compression then comes in. Using the chosen eigenvectors the dimension of the data can be reduced while retaining a large amount of information. By using only the largest eigenvalues and their corresponding eigenvectors compression can be used as well as a simple transformation. Thus the data compression or transformation is (5):

$$P = D \times PC \quad (5)$$

Where $D$ is the original data set, PC is the principal component matrix and $P$ is the transformed data. The principal component analysis multiplication results in a data set that emphasises the relationships between the data

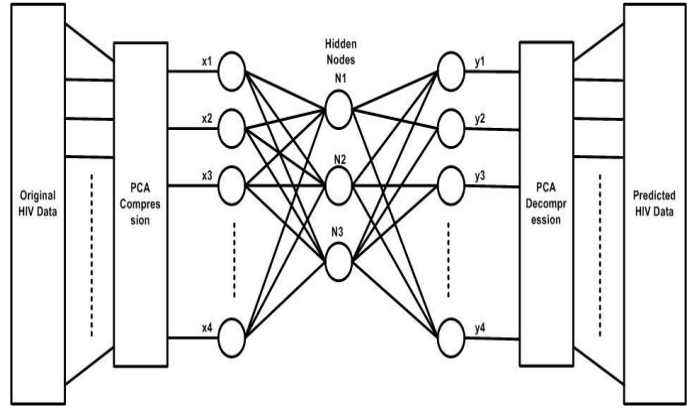

whether smaller or the same dimension. To return to the original data the following equation is used (6):

$$D' = P \times PC^{-1} \quad (6)$$

Here $D'$ is the retransformed data. If all of the principal components are used from the covariance matrix then $D = D'$. The transformed data ($D$) can be used in conjunction with the ANN to increase the efficiency of the ANN by reducing its complexity (number of training cycles). These results from the property of the PCA extracting linear relationships between the data variables, thus the ANN only needs to extract the non linear relationships. This then results in less training cycles that are needed. Thus ANNs can be built more efficiently. Fig. 3 illustrates this concept. The PCA function in Netlab was used for the investigation 0.

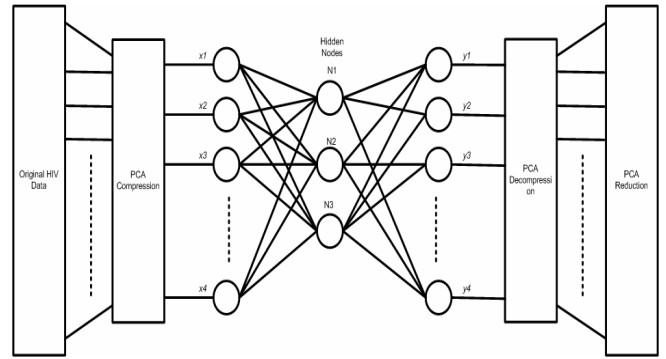

Fig. 3. PCA Autoencoder Neural Network

*2.6 Genetic Algorithms*

Genetic algorithms are defined as population based models that use selection and recombination operators to generate new sample points in search space (Whitley, 1994). Genetic algorithms are primarily used for optimisation as they can find values for variables that will achieve a target. In this investigation the genetic algorithm is used to find the input into regression model that will result in the most accurate missing data value. Genetic algorithm use is good for non linear functions and applications, thus the use in this investigation. The overview of the procedure of genetic algorithm is the same as that of natural selection.

A genetic algorithm starts with a creation of a random population of "chromosomes". These chromosomes are normally in binary format. From this random population an evaluation function is used to find which of the chromosomes is the fittest. Those who are deemed fit are then used for the selection stage. Recombination of the chromosomes is done by taking the fittest chromosomes and choosing bits from each that will be swapped (deemed crossover). This then results in a new population of chromosomes. The final stage is mutation were bits are then randomly changed within the chromosomes. From this new population the fitness operation begins again until a preset number of iterations. The genetic algorithm toolbox was used for the investigation (Houck, 1995).

### 3. DATA COLLECTION AND PRE-PROCESSING

The data that is used for this investigation is HIV data from antenatal clinics from around South Africa. It was collected by the department of health in the year 2000. The data contains multiple input fields that result from a survey. The information is in a number of different formats resulting from the survey. For example the provinces, region and race are strings. The age, gravidity, parity etc. are integers. Thus conversions are needed. The strings were converted to integers by using a lookup table e.g. there are only 9 provinces so 1 was substituted for Gauteng etc.

Data collected from surveys and other data collection methods normally have outliers. These are normally removed from the data set. In this investigation data sets that had outliers had only the outlier removed and the data set was then classified as incomplete. This then means that the data can still be used in the final survey results if the missing values are imputed. The data with missing values was not used for the training of the computational methods. The data variables and their ranges are shown below in Table 1.

Table 1. HIV Data Variables

| Variable | Type | Range |
|---|---|---|
| HIV Status | Binary | [0, 1] |
| Education | Integer | 0 - 13 |
| Age Group | Integer | 14 - 60 |
| Age Gap | Integer | 1 - 7 |
| Gravidity | Integer | 0 - 11 |
| Parity | Integer | 0 - 40 |
| Race | Integer | 1 - 5 |
| Province | Integer | 1 - 9 |
| Region | Integer | 1 - 36 |
| RPR | Integer | 0 - 2 |
| WTREV | Continuous | 0.638 – 1.2743 |

The pre-processed data resulted in a reduction of training data. This was 12750 processed data sets from around 16500 original records in the survey data. To use the data for training it needs to be normalised. This ensures that the all data variables can be used in training. If the data is not normalised, some of the data variables with larger variances will influence the result more than others. E.g. if we use WTREV and Age Group data only the age data will be influential as it has large values. Thus all of the data is normalised between 0 and 1. The training data is then split into 3 partitions. 60% is used for training, 15% for validation and the last 25% used for the testing stages.

### 4. METHODOLOGY

The approach taken for the project is to use the regression methods with an optimisation technique. The optimisation technique chosen was the Genetic algorithm. Fig. 4 illustrates the manner in which the regression methods and the optimisation technique will be used to impute data

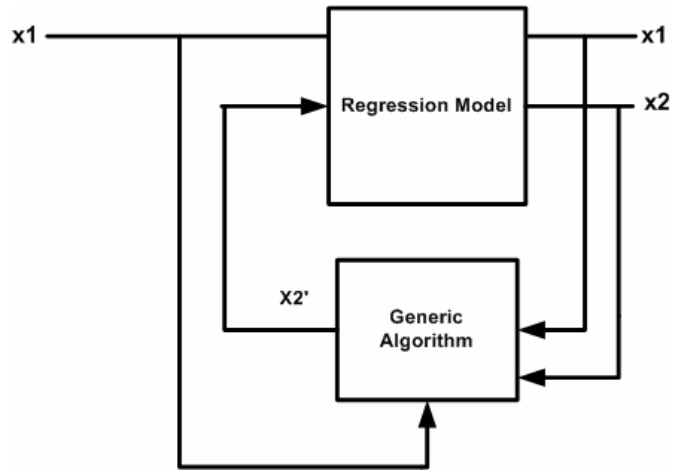

Fig. 4. Data imputation Configuration

The regression methods first had to be trained before being used for data imputation. The following subsections discuss the training procedures for the regression methods.

*4.1 ANN Training and Validation*

To train the ANN the optimum number of hidden nodes is needed. To find it a simulation was constructed that calculated the average error using a different number of hidden nodes. The number of hidden nodes were optimised, and found to be 10. This was using scaled conjugate gradient a linear outer activation function and a hyperbolic tangent function as the inner activation function. Then the optimal number of training cycles needed to be found. This was done by analysing the validation error as the training cycles increased. This is to both avoid the possibility of overtraining the ANN and use the fastest way to train the ANN without compromising on accuracy. It was found that 1000 training cycles were sufficient as well as the use of the early stopping method if the ANN was beginning to be over trained. Validation was done with a data set that was not used for training. This then resulted in an unbiased error check that would indicate if the network was well trained or not.

*4.2 PCA ANN Training and Validation*

The training data was first used to extract the principal components. After the extraction the training data was multiplied with the principal components and the resulting data was used to train a new ANN. This was then labelled a PCA-ANN. Two PCA-ANNs were trained. One PCA-ANN had no compression and was just a transform; the other

PCANN compressed the data from 11 dimensions to 10. The number of hidden nodes and training cycles were optimised as in the previous subsection. The number of hidden nodes for the PCA-ANN-11 was 10 and for the PCA-ANN-10 were 9. The inner and outer activation functions were as for the ANN above. Validation was also carried out with an unseen data set. This also ensures that the ANN is trained well and not over trained.

*4.3 SVR Training and Validation*

To train the support vector regression model less training data was needed. Only 3000 data records were used in this case, this was due to time constraints, the training took a considerable amount of time on MATLAB. Even though a smaller training set was used the validation error was small. A radial basis function kernel function was used. The bias point and the regularisation had to be optimised. To optimise the two a genetic algorithm was utilised. This technique has been used by Kuan-Yu Chen and Chen-Hua Wang (Chen, 2007) with good results. The GA used a validation set to find the parameters that resulted in the minimum error in a SVR regression validation data set. Validation was carried out after training with an unseen set and the SVR performed well. This was satisfactory and the SVR could now be used with the GA to impute missing data.

*4.4 Genetic Algorithm Configuration*

The genetic algorithm will be configured as in the model in Fig. 5.

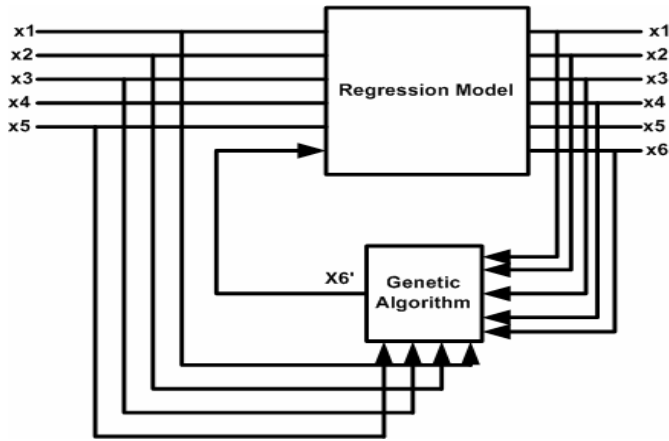

Fig. 5. Genetic Algorithm Configuration

The inputs $\{x_1 - x_5\}$ are known, $x_6$ is unknown and will be found by using the regression method and the GA. The genetic algorithm will put a value from its initial population into the regression model. The model will recall the value and it will be an output. The GA will try and minimise the error between its approximated value and the value that the regression model will have as an output. This will be done via the fitness function of the genetic algorithm. The fitness evaluation function of a GA is normally a reduction of error function such as in (7):

$$e = (x - y) \quad (7)$$

Where $x$ is the known value and $y$ is the estimated value. As the GA locates a global maximum and not a minimum equation 3 has to be changed to the form of (8):

$$e = -(x - y)^2 \quad (8)$$

The error thus approaches zero from the negative axis and thus the GA will be able to find a global maximum. In this project the genetic algorithm uses normalised geometric selection for selection, along with simple crossover for recombination and non uniform mutation. The GA is used to approximate the values of the missing data and using the auto encoder network or the SVR mechanism then uses the evaluation function (9) below to calculate the fitness.

$$e = -\left(\begin{Bmatrix} x_u \\ x_k \end{Bmatrix} - f\left(\begin{Bmatrix} x_u \\ x_k \end{Bmatrix}\right)\right)^2 \quad (9)$$

As ANN and the SVR are used they try to recall their inputs. $x_u$ is the unknown parameter that has just been approximated by the GA while $x_k$ are the known data variables as $x_1 - x_5$ in Fig. 5. The function $f$ is the regression model and is changed for each of the models previously discussed. Fig. 5 shows the configuration of the GA with the regression methods. The GA will try and reduce the error between the regression method and the data inputs. Thus resulting in a data variable that is likely to be the missing value. But for completeness all of the outputs are used to reduce the error of the approximated value. The regression methods discussed in the preceding section were combined with the Genetic algorithm as in Fig. 5. This results in multiple data imputation mechanisms. These are the:

- ANNGA (combination of the ANN and GA)
- PCANNGA (combination of PCA, ANN and GA)
- SVRGA (combination of SVR and GA).

The Genetic Algorithm was setup with 50 initial population and 50 generation cycles. As mentioned earlier the GA uses simple crossover, geometric selection and non uniform mutation. This produced the best results and was used for every model so as to serve for correct comparisons.

## 5. TESTING

The testing set for the data imputation methods contained 1000 sets. These were complete data sets that had some of their data removed so as to be able to ascertain the accuracy of the imputation methods. The testing set is made up of data that the imputation methods have not seen yet (i.e. Data that is not part of the training or validation set). This data was also chosen randomly from the initial dataset that is outlined in Section 3. The variables to be imputed where chosen to be HIV status, Age, Age Group, Parity and Gravidity. These were taken as the most important data variables the needed to be imputed. The testing sets were comprised of 3 different data sets made up of a 1000 random records each. This offers an unbiased result as testing with only 1 test can have results which are the best but may be biased due to the data used.

*5.1 Methods for measuring accuracy*

Different measures of accuracy are used for the evaluating the effectiveness of the imputation methods. This is to offer better understanding of the results. The accuracy measures are discussed below.

*5.2 Mean Square Error*

The mean square error is used for the regression and classification data. It is used to measure the error between the imputed data and the real value data. It is expressed as (10):

$$e = (x-y)^2 / n \quad (10)$$

*x* is the correct value data and y is the imputed data. n is the number of records in the data. The mean square error is calculated after the imputation by the GA. This is before de-normalisation and rounding. Thus does not carry over any rounding errors.

*5.3 Classification Accuracy*

For the classification value of the HIV data the only accuracy used is the number of correct hits. This means the number of times the imputer imputed the correct status. This is done after de-normalisation and rounding.

*5.4 Prediction within Years/Unit Accuracy*

Prediction within year is used as a useful and easy to understand measure of accuracy. This for example would be expressed as 80% accuracy within 1 year for age data. This means for age data the values that are found are 80% accurate within a tolerance of 1 year. This measure is used mainly for the some of the regression data.

## 6. RESULTS

All of the results shown in the tables are in percentages of accuracy. For HIV, Gravidity, Parity and Age Gap are positive match accuracy. Education Level and Age are all accuracies with a tolerance of 1 year.

*6.1 ANNGA*

The ANNGA was tested with all of its optimised variables and trained network. The results of the ANNGA data imputation are tabulated in Table 2.

Table 2. ANNGA Results

| ANNGA(%) | Run 1 | Run 2 | Run 3 | Average |
|---|---|---|---|---|
| **HIV Classification** | 68.9 | 68.6 | 68.0 | 68.5 |
| **Education Level** | 25.1 | 25.1 | 27.2 | 25.8 |
| **Gravidity** | 82.7 | 82.0 | 84.0 | 82.9 |
| **Parity** | 81.3 | 81.1 | 82.1 | 81.5 |
| **Age** | 86.9 | 86.4 | 85.5 | 82.3 |
| **Age Gap** | 96.6 | 96.0 | 95.4 | 96 |

The results indicate that the autoencoder network genetic algorithm architecture seems to perform well in the HIV classification and as well all the others except the education level. The high estimation accuracies are on par with previous research. The education level seems to be the weak point.

*6.2 PCANNGA*

The PCANNGA architecture was run with two configurations. The first configuration had no compression thus is named PCANNGA11 indicating the transformation from 11 inputs to 11 outputs. The second configuration has a compression of 1 value thus is named PCANNGA-10, indicating the compression and transformation from 11 inputs to 10 inputs. The results of the test are shown below in Table 3.

Table 3. PCANNGA Results

| PCANNGA–11 (%) | Run 1 | Run 2 | Run 3 | Average |
|---|---|---|---|---|
| **HIV Classification** | 65.0 | 61.6 | 62.8 | 63.1 |
| **Education Level** | 27.8 | 27.3 | 28.2 | 27.8 |
| **Gravidity** | 87.6 | 86.5 | 87.1 | 87.1 |
| **Parity** | 87.5 | 86.3 | 87.7 | 87.2 |
| **Age** | 94.9 | 94.8 | 93.5 | 95.7 |
| **Age Gap** | 98.1 | 98.3 | 96.9 | 97.4 |
| PCANNGA –10(%) | Run 1 | Run 2 | Run 3 | Average |
| **HIV Classification** | 64.2 | 60.9 | 67.2 | 64.1 |
| **Education Level** | 27.0 | 31.3 | 30.2 | 29.5 |
| **Gravidity** | 86.4 | 86.3 | 88.2 | 61.0 |
| **Parity** | 86.2 | 86.2 | 87.6 | 86.7 |
| **Age** | 8.0 | 8.2 | 12.1 | 9.4 |
| **Age Gap** | 23.9 | 20.0 | 24.1 | 22.7 |

The results for PCANNGA-11 indicate good estimation for all the variables except education level. PCANNGA-10 performs poorly on Age and Age Gap while having good results in the other variables. This results from the loss of information during the compression. This then impacts on the regression ability of the network resulting in poor imputation accuracy for some of the variables.

*6.3 SVRGA*

The SVRGA imputation model took a long time to run. Due to the inefficiencies of running a computational such as this on MATLAB, the simulations were slow. Nonetheless the imputations did run and did return all required results. The results from the SVRGA are tabulated below in Table 4.

Table 4. SVRGA Results

| SVRGA (%) | Run 1 | Run 2 | Run 3 | Average |
|---|---|---|---|---|
| **HIV Classification** | 22.5 | 22.1 | 21.4 | 22 |
| **Education Level** | 65.4 | 40.3 | 45.6 | 50.433 |
| **Gravidity** | 80.9 | 63.2 | 67.4 | 70.5 |
| **Parity** | 81.4 | 63.3 | 66.9 | 70.5 |
| **Age** | 96.1 | 89.2 | 83.5 | 89.6 |
| **Age Gap** | 92.6 | 92.7 | 94.3 | 93.2 |

The SVRGA performs badly in the HIV classification. It performs averagely in the Education level, Parity and Gravidity. With Age and Age gap it performs well

*6.4 Comparison of Results*

For the comparison of results, the previous accuracies as well as the mean square error of each method will be analysed. This will give an indication of how the errors in the imputation affect the accuracy as well as which model produces the best results. The average mean square errors of the imputation methods are shown in Table 5

Table 5. Average Mean Square Errors

| Results | NN | PCANN11 | PCANN10 | SVRGA |
|---|---|---|---|---|
| **HIV** | 0.269147 | 0.303703 | 0.301647 | 0.764407 |
| **Education** | 0.16663 | 0.13224 | 0.123517 | 0.0421 |
| **Gravidity** | 0.00187 | 0.001456 | 0.001478 | 0.003141 |
| **Parity** | 0.002592 | 0.00237 | 0.002373 | 0.004422 |
| **Age** | 0.001025 | 0.000396 | 0.157397 | 0.003178 |
| **Age Gap** | 0.001289 | 0.000548 | 0.097913 | 0.002087 |

In the mean square errors a smaller value is desirable. It can be seen from Table 5 that in HIV classification the SVRGA performed the worst as it had the highest error but in the education level it performed the best as it has the lowest error. The following figure, Fig. 6, is a graph of the average mean square error of the imputation models

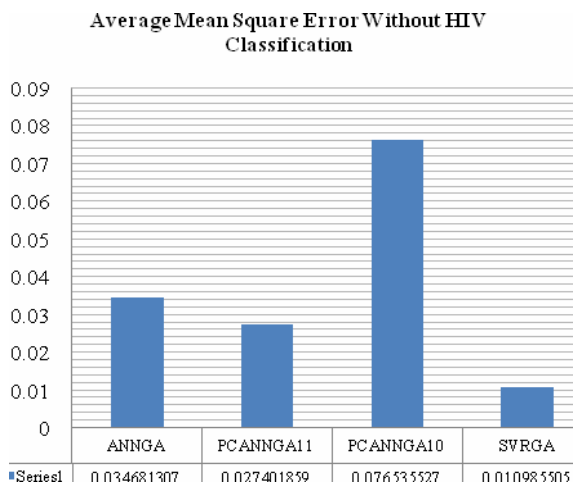

Fig. 6. Comparison of Average Mean square error without HIV Classification

From Fig. 6 it can be seen that the SVRGA has the smallest average mean square error (if HIV classification is not included) from the rest of the methods. This indicates that the SVRGA functioned well on regression parameters and poorly on the classification of HIV. The following graph in Fig. 7. makes this clear. The ANNGA performs the best with an average accuracy of 68.5 % while the rest of the models fell behind and the SVRGA has the lowest average accuracy of 22 %. In Education level accuracy the SVRGA performed

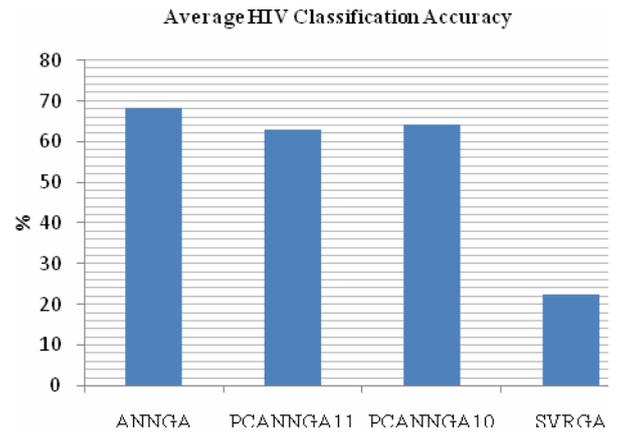

best. It had an overall accuracy of 50%. This is measured

Fig. 7. Comparison of HIV Classification Accuracy

within a tolerance of 1 year. The accuracies of the models are shown in Fig. 8.

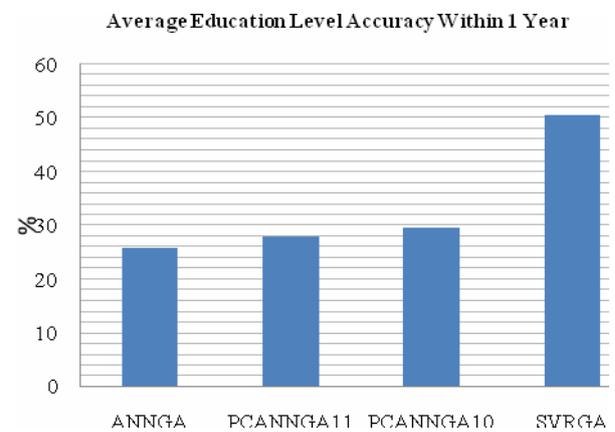

Fig. 8. Comparison of Average Education Accuracy

The SVR is predicting the education level better than the rest and thus is performing better when combined with the Genetic Algorithm to impute the missing variables. The last comparison is of the age accuracy. The average accuracies with 1 year tolerance are shown in Fig. 9. From the graphs it can be seen that the PCANN10 overall performs poorly. As explained earlier this results from the data loss from the compression of the data. The SVRGA performs better than the ANNGA but the PCANN11 performs better than all. In almost all of the accuracy tests the PCANN11 performs better than the ANNGA thus proving that the combination of the PCA and ANN can result in a better imputation method. The PCANN11 even has a lower average mean square error than the ANNGA as shown in Fig. 6. The PCA without compression has improved the performance of the ANNGA.

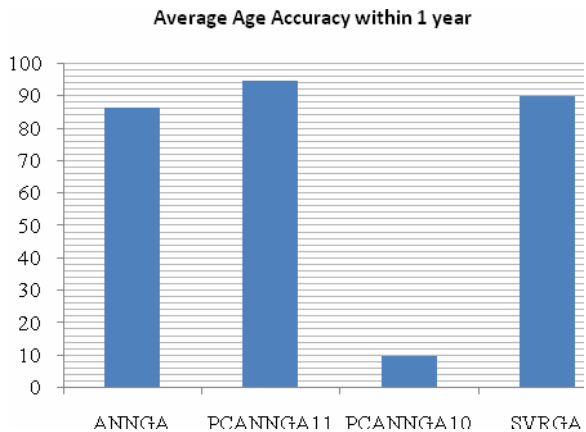

Fig. 9. Comparison of Age Accuracy

From the comparison of all of the imputation models it can be seen that the PCANN11 performs better even though it has a worse HIV classification. The SVRGA only makes good ground on the education level and thus cannot be considered superior to the PCANN11

## 7. DISCUSSION

### 7.1 General Performance

The general performance of the imputation methods is satisfactory and highly accurate. The high accuracy of the imputation methods on the variables makes them a viable solution for the Department of Health's HIV/AIDS research. This affords researchers confidence that the data collected does not have to have a large amount of it discarded. The ANNGA neural network is stable and the results were good. The SVRGA performed the best in the education level and this could be further investigated. The PCANNGA11 on average shows the best promise in high accuracy missing data imputation. This results from its good average performance in Parity, Gravidity, Age and Age Gap while only lagging behind by a small margin in the HIV classification and performing better than the ANNGA in predicting the Education Level. Solutions with higher tolerances tend to be given but the low tolerance used in this investigation was to illustrate the high accuracies. Higher tolerances can be used selectively and instead of years in a variable like education levels can be put into 3 categories like primary school, high school and tertiary. This has been done by with rough set theory in *(*Nelwamondo *et al,* 2007b*)*.

### 7.2 SVRGA

Due to time constraints the support vector regression could not be investigated further. This is due to the fact that the simulations of the SVRGA were very slow. SVR though is still a viable solution if an optimised c++ or other programming language toolbox is used instead of a MATLAB toolbox, the speed of computation will increase. Thus it is suggested that more research and investigation be done on the SVR. There have been cases were the SVR has outperformed normal neural networks. Thus the author believes a SVR can outperform an ANN.

### 7.3 Hybrid

A hybrid approach of using the ANNGA and SVRGA or PCANNGA11 and SVRGA together is also a viable future investigation area. This could not be implemented in the investigation due to time. It is expected that this would increase the performance of the neural network based methods in imputing the education level while assisting the SVRGA in imputing the HIV classification.

### 7.4 Further Regression vs. Classification

An investigation into the data only for classification for the classification parameters such as HIV can yield better results. This comes at the price of loss of generalisation. Leke and Marwala (Leke *et al*, 2005) investigated a classification based problem of HIV classification only. This cannot be directly used with data imputation without then resulting in high complexity hybrid networks with models only dealing with missing data that is classification based and then other models dealing with regression based missing data.

## 8. CONCLUSION

This paper investigated and compared the use of 3 regression methods with a GA combination for missing data approximation. An autoencoder neural network was trained to predict its input space, as well as reconfigured with a principal component analysis to form a principal component analysis autoencoder neural network that predicted the principal component transformed input space. Support vector regression was also used in the same manner as the autoencoder network. The regression methods were combined with genetic algorithms to approximate missing data from a 2000 HIV survey data set. The principal component autoencoder neural network genetic algorithm model performed the best overall, with accuracies up to 97.4%, followed by the autoencoder neural network genetic algorithm model. The support vector regression genetic algorithm model performed well on approximating a missing variable that the rest of the models performed poorly in. This allows for future investigation into hybrid systems with combinations of the regression models in order to get better results and better methods for future data imputation.